\documentclass[manuscript]{copernicus}

\makeatletter
\AtBeginDocument{%
  \nolinenumbers
  \let\linenumbers\nolinenumbers
}
\makeatother

\usepackage{booktabs}

\begin{document}

\title{LUCAS-MEGA: A Large-Scale Multimodal Dataset for Representation Learning in Soil--Environment Systems
}


\Author[1]{Kuangdai}{Leng}
\Author[1,2]{Simon}{Jeffery}
\Author[3]{Panos}{Panagos}
\Author[1,4]{Tarje}{Nissen-Meyer}

\affil[1]{Earth Rover Program, 71-75 Shelton Street, London WC2H 9JQ, United Kingdom}
\affil[2]{Centre for Crop and Environmental Science, Harper Adams University, Newport, Shropshire TF10 8NB, United Kingdom}
\affil[3]{European Commission Joint Research Centre, Via E. Fermi, 2749, 21027 Ispra VA, Italy}
\affil[4]{Department of Mathematics and Statistics \& Center for Environmental Intelligence, University of Exeter, Exeter EX4 4QF, United Kingdom}




\correspondence{Tarje Nissen-Meyer (tarje@earthroverprogram.org)}

\runningtitle{LUCAS-MEGA}

\runningauthor{Leng et al.}

\received{}
\pubdiscuss{} 
\revised{}
\accepted{}
\published{}


\firstpage{1}

\maketitle

\begin{abstract}
Understanding soil is fundamental to agriculture, carbon cycling, and environmental sustainability, yet progress is limited by fragmented and heterogeneous datasets that constrain modeling to small-scale predictive settings rather than high-dimensional representation learning. We introduce LUCAS-MEGA, a large-scale multimodal dataset constructed through systematic data fusion of European soil--environment observations, with the LUCAS survey as its backbone. The fused dataset comprises over 70,000 samples and more than 1,000 features spanning physical, chemical, environmental, biological, and visual attributes, aggregated from 68 source datasets. To enable integration at scale, we develop SoilFuser, a multi-agent, human-in-the-loop data fusion pipeline that standardizes heterogeneous data formats and measurement protocols, resolves inconsistencies and invalid entries (e.g., unit inconsistencies, codebook mismatches, and erroneous values), incorporates natural language annotations, and harmonizes multimodal attributes and metadata into a unified, machine learning-ready feature space. The resulting dataset captures key characteristics of real-world soil observations, including multimodality, uneven feature coverage, and heterogeneous uncertainty. To demonstrate the usability of LUCAS-MEGA for data-driven modeling, we pretrain a multimodal tabular transformer (SoilFormer) using a self-supervised objective based on feature masking, achieving stable training, strong predictive performance, and representations that support uncertainty-aware prediction. We further show that the learned representations recover relationships consistent with established soil processes. LUCAS-MEGA is released with open access and is accompanied by composable, agent-friendly APIs that support structured querying and data-driven workflows.
\end{abstract}


\section{Introduction}

Soils are increasingly recognized as a critical but threatened foundation of terrestrial ecosystems, agricultural production, climate regulation, and biodiversity. In Europe, for example, 62\% of soils are estimated to be unhealthy based on the 19 soil degradation indicators of the EU Soil Health Dashboard~\citep{panagos2024euso}. This urgency is further reflected in the EU Soil Monitoring Law, which establishes a harmonized framework for monitoring soil health across multiple soil threats and descriptors, reflecting the need to assess soil properties and degradation processes in an integrated rather than isolated manner~\citep{panagos2025soilmonitoring}. Such interdependence reflects the inherent complexity of soils: their states and behaviors emerge from tightly coupled physical, chemical, and biological processes interacting across spatial and temporal scales~\citep{jenny1941factors,young2004selforganization,turner2021complexity}. Capturing this complexity therefore requires jointly modeling a large number of interdependent variables, often with uneven coverage across regions, depths, and observation systems, resulting in \textit{a high-dimensional and heterogeneous feature space}.

Recent advances in scientific machine learning have led to more general-purpose and multimodal foundation models~\citep{vaswani2017attention,bommasani2021opportunities}. These models learn shared representations across large numbers of variables and modalities~\citep{lam2023graphcast,jakubik2025terramind}, enabling the discovery of complex cross-variable relationships and supporting a variety of downstream tasks. This paradigm is well aligned with the need to model soil as an integrated system, but its application in soil science remains constrained by the lack of unified, high-dimensional, and machine learning-ready data. Existing sample-based soil data are fragmented across sources, heterogeneous in format and measurement protocols, and not organized to support joint modeling across variables and modalities. As a result, most existing data-driven approaches in soil science rely primarily on small-scale, task-specific models trained on limited subsets of variables, providing only a partial view of soil--environment systems; see reviews by \cite{minasny2024soil,minasny2025machine}. 

Soil data with broad geographic coverage are available through a range of spatial soil resources, from expert-derived soil databases such as the European Soil Database (ESDB)~\citep{king1994esdb} and the Harmonized World Soil Database (HWSD)~\citep{faoiiasa2012hwsd}, to digital soil mapping (DSM) products such as SoilGrids~\citep{poggio2021soilgrids2}. These products provide spatially continuous estimates at resolutions ranging from hundreds of meters to kilometers. While highly valuable for sub-national to continental-scale assessment and policy applications, they are typically derived from field observations and environmental covariates through empirical knowledge, statistical interpolation, or machine learning models, and thus constitute spatially generalized representations rather than direct measurements. This introduces inductive biases into the resulting products and limits their ability to capture fine-scale variability driven by local environmental and management factors. In addition, these products are limited in their feature space, providing only a small set of key soil properties and lacking the variable breadth needed for integrated representation learning.

Direct soil observations are available from large-scale field surveys, including topsoil monitoring programs such as LUCAS (Europe)~\citep{orgiazzi2018lucas}, RMQS (France)~\citep{armand2018rmqs}, and NSI (UK)~\citep{lark2012nsi}, as well as soil profile-based datasets such as WoSIS (Global)~\citep{batjes2017wosis}, SPADE/M (Europe)~\citep{hiederer2006spadem}, EU-HYDI (Europe)~\citep{weynants2013euhydi}, AfSIS (Africa)~\citep{hengl2015afsis}, and NCSS (US)~\citep{ncss2023}. These datasets provide sample-based, directly measured observations and are therefore well suited for data-driven modeling. However, they originate from heterogeneous sources with differing measurement protocols, data formats, and variable coverage, and are not organized into a unified representation that supports joint modeling across datasets. In addition, the feature space of individual datasets remains limited: each sample typically contains tens of measured variables, reflecting the cost and design constraints of field and laboratory measurements.

Recent work has increasingly explored multimodal and multi-source datasets or benchmarks in environmental and agricultural science, often incorporating soil information as part of a broader modeling context. Examples include IRRISIGHT for irrigation mapping and agricultural water management~\citep{mandal2025irrisight}; CY-Bench, CYCleSS, and a global conventional-tillage/no-tillage dataset for crop-yield modeling~\citep{paudel2025cybench,corcoran2026cycless,su2021cropctnt}; AgriBench/MM-LUCAS and AgMMU for benchmarking multimodal or vision--language models in agriculture~\citep{zhou2025agribench,gauba2025agmmu}; and SoilNet for multimodal hierarchical classification of soil horizons~\citep{chiaburu2025soilnet}. These efforts demonstrate the growing importance of multimodal and multi-source data in agriculture and environmental modeling. However, where soil information is included, it is typically represented by a small set of core attributes, such as soil type, texture, bulk density, or selected chemical properties, and serves mainly as task-specific context rather than as a basis for soil-centered representation learning across high-dimensional soil--environment features.

Collectively, existing soil datasets are either spatially comprehensive but low-dimensional and model-derived, or observation-based but fragmented and heterogeneous. There remains a lack of datasets that support high-dimensional representation learning of soil systems, where a large number of interdependent variables with uneven coverage are modeled jointly to capture the intrinsic complexity of soils. Related efforts in other data-intensive scientific domains provide useful precedents, such as FLUXNET~\citep{pastorello2020fluxnet2015} for environmental monitoring networks and MIMIC~\citep{johnson2023mimiciv} for healthcare.

\begin{figure}
    \centering
    \includegraphics[width=0.75\linewidth]{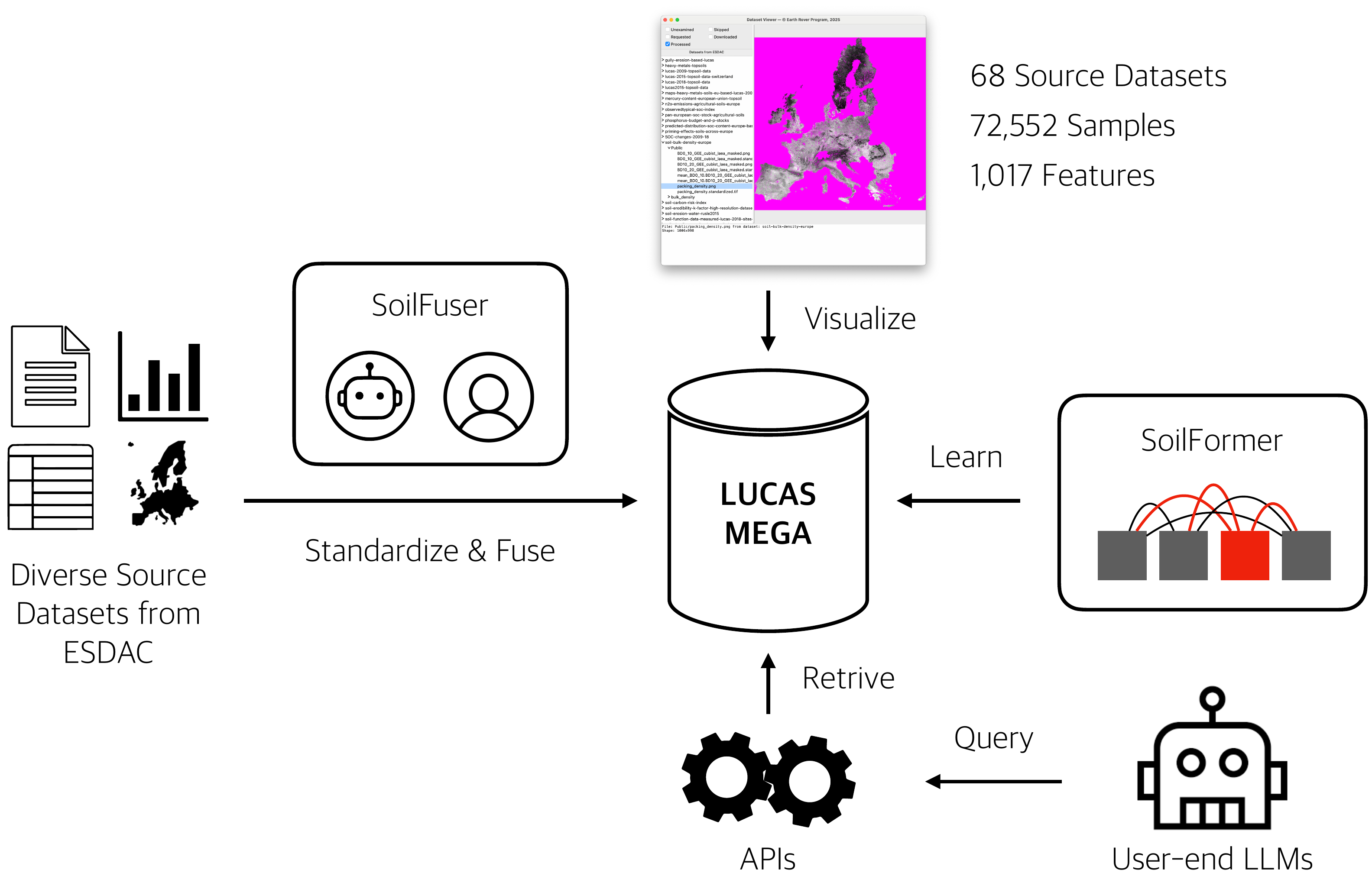}
    \caption{Overview of the LUCAS-MEGA pipeline from data fusion to modeling and application.}
    \label{fig:pipeline}
\end{figure}

To address these gaps in soil science, we introduce \textbf{LUCAS-MEGA}, a large-scale multimodal dataset for soil systems constructed through systematic data fusion of European soil--environment observations. Our dataset uses the LUCAS soil survey as its primary backbone, augmented with additional sources including SPADE/M and EU-HYDI, resulting in \textbf{a unified sample space of over 70,000 entries}. To enable feature integration across heterogeneous sources, we develop \textbf{SoilFuser}, a multi-agent, human-in-the-loop data fusion system that standardizes and harmonizes disparate datasets into a unified feature space. Using this system, we analyze and curate 130 independent datasets hosted by the European Soil Data Centre \citep[ESDAC; as of September 2025;][]{panagos2022esdac}, incorporating 68 of them into LUCAS-MEGA and constructing \textbf{a high-dimensional feature space with over 1,000 features}. Figure~\ref{fig:pipeline} provides a high-level overview of the LUCAS-MEGA system, including data fusion, modeling, and access.

Beyond its scale, LUCAS-MEGA is characterized by the following properties:

\begin{itemize}
    \item {(\textbf{M})}ulti-modal -- Features span numerical (scalar and vector-valued), categorical, textual, and visual modalities, covering physical, chemical, biological, environmental, functional, and soil threat-related attributes.
    
    \item {(\textbf{E})}nd-to-end machine learning-ready (ML-ready) -- The dataset is provided as a unified, schema-aligned representation with standardized units, consistent formats, and machine-readable metadata, enabling direct use in data-driven and machine learning workflows.
    
    \item {(\textbf{G})}reat quality -- Extensive validation and correction are performed to resolve data quality issues such as unit inconsistencies, invalid values, inconsistent missing-value codes, and ambiguous or outdated codebook definitions.
    
    \item {(\textbf{A})}ccessible -- LUCAS-MEGA is released with open access, accompanied by visualization tools and retrieval APIs that support both data exploration and model development.
\end{itemize}

The main objective of LUCAS-MEGA is to bridge the gap between fragmented European soil--environment data resources and the requirements of modern data-driven soil science. First, for soil scientists and environmental practitioners, it provides a centralized resource with improved consistency and quality relative to the original raw datasets, thereby facilitating data discovery, integration, and reuse. Second, for soil-centered machine learning, it provides a large-scale, multimodal, and partially observed dataset for modeling variables spanning soil properties, environmental covariates, and soil threats, supporting tasks such as prediction, data gap filling, and assessing management interventions. Finally, beyond soil science, LUCAS-MEGA provides a structured case study for representation learning from high-dimensional, heterogeneous scientific data with uneven feature coverage, structured missingness, cross-variable dependence, and measurement uncertainty; this setting can help evaluate methods that aim to learn coherent and uncertainty-aware representations from complex real-world datasets.

To demonstrate the utility of LUCAS-MEGA, we conduct experiments on multimodal representation learning for soil systems. We pretrain a multimodal tabular transformer, \textbf{SoilFormer}, using a BERT-style self-supervised objective based on feature masking~\citep{devlin2019bert}, enabling joint modeling across high-dimensional and heterogeneous variables. Our results show stable training at scale and strong predictive performance across multiple soil properties, while the learned representations reflect inherent measurement noise through heteroscedastic (aleatoric) uncertainty learning~\citep{kendall2017uncertainties} and recover relationships consistent with established soil processes. These findings highlight the effectiveness of LUCAS-MEGA as a reliable resource for integrated soil modeling. In parallel, the composable APIs and prompt-based interfaces built around LUCAS-MEGA enable retrieval-augmented, data-grounded reasoning with user-facing large language models, supporting both analytical and interactive workflows.

The remainder of this paper is organized as follows. Section~\ref{sec:data_construction} describes the data sources and the dataset construction process, including our agent-driven pipeline, SoilFuser. Section~\ref{sec:dataset} presents the LUCAS-MEGA dataset, detailing its sample structure, feature space, modality coverage, and key statistical characteristics. As downstream applications, Section~\ref{sec:rag} introduces data access interfaces for retrieval-augmented workflows, while Section~\ref{sec:learn} demonstrates the use of LUCAS-MEGA for multimodal representation learning.

\section{Data Construction and Integration}
\label{sec:data_construction}

\subsection{Data Sources}
\label{sec:data_sources}

We consider the full collection of \textbf{130 datasets} hosted by ESDAC as the source pool for data fusion. ESDAC is selected for three main reasons:
\begin{enumerate}
    \item It offers both centralization and breadth, hosting a large number of datasets spanning diverse domains, including soil properties, environmental covariates, land use and management, and geospatial products.
    \item The datasets exhibit substantial heterogeneity in data formats (e.g., GeoTIFF, CSV, Excel, text, and binary), spatial resolution, data volume, and documentation quality, providing a realistic and challenging setting for developing a robust general-purpose data fusion pipeline. Developing the pipeline on ESDAC therefore helps ensure applicability to other data hosts with similar diversity and complexity.
    \item ESDAC hosts the LUCAS soil survey~\citep{orgiazzi2018lucas}, a cross-country, multi-year topsoil survey with standardized sampling and measurement protocols across Europe. LUCAS provides consistent and georeferenced observations at scale, forming the basis for constructing a sample-specific, ML-ready dataset.
\end{enumerate}

\subsection{Data Standardization and Fusion Pipeline}
\label{sec:soilfuser}

All the 130 datasets downloaded from ESDAC undergo three steps: screening, standardization and fusion.

\subsubsection{Screening Criteria}

The 130 datasets are manually reviewed prior to processing. We exclude datasets that (i) are not directly related to soil or soil--environment processes (e.g., policy-related datasets or auxiliary tools), (ii) are aggregated at coarse administrative levels (e.g., national or regional statistics), (iii) are fully covered by other datasets (e.g., rasters derived from vector databases), or (iv) are excessively large (on the order of tens to hundreds of gigabytes) with limited relevance to soil–environment characterization.

This screening step ensures both relevance and diversity of the retained datasets while avoiding redundancy and impractical data volumes. After screening, \textbf{96 datasets} are retained as candidates for data fusion.

\subsubsection{Standardization Scheme}
\label{sec:standardization}

Despite differences in file formats and data organization, the retained raw data can be grouped into two structural types: \textit{sample-structured} and \textit{map-structured}, with some datasets containing both. Sample-structured data consist of georeferenced records associated with point locations, which will directly map to soil observations in our unified representation while contributing feature values. Map-structured data are defined on grids or polygons, which will be queried at sample locations to provide additional features, with resolution-induced uncertainty determined by spatial distance. In most cases, sample-structured data correspond to field measurements, while map-structured data are derived from statistical or machine learning models.

All map-structured data are standardized into GeoTIFF format with a unified coordinate reference system and explicit no-data masking. Rendered image files are also generated for rapid inspection.
All sample-structured data are standardized into CSV format with unified metadata, with particular attention to the following:
\begin{enumerate}
    \item Categorical variables are harmonized by resolving and standardizing codebooks. In many cases, categorical labels are not directly interpretable or inconsistently documented, requiring cross-referencing with external source materials. These codebook-based mappings are curated to ensure semantic consistency across datasets. 
    \item Outlier detection is performed on numerical properties via out-of-distribution analyses; for example, values such as pH and bulk density reported as 0.0 are identified as invalid placeholders for missing measurements.
    \item High-dimensional vector-valued properties (e.g., particle size distributions from LUCAS and water retention curves from EU-HYDI) are stored as ``assets'' and linked to samples. 
\end{enumerate}

Based on these standardized representations, a lightweight viewer is developed to visualize data and metadata in a unified interface (Figure~\ref{fig:pipeline}). Although fusion is not performed at this stage, this standardized collection of datasets already provides a consistent and accessible representation, offering value in addition to the final fused dataset.

\subsubsection{Fusion Scheme}

The target fused dataset is organized in a sample-based representation, where each sample corresponds to a soil observation associated with a georeferenced location and is linked to a set of features aggregated from multiple sources, forming an ML-ready configuration. Together with the standardized representations obtained in Section~\ref{sec:standardization}, the abstractions of both the input and output of the fusion process become well defined.

We therefore adopt a schema-driven approach, in which each dataset is associated with a schema that specifies how its records are mapped into the unified target representation. The schema defines feature correspondence and cross-dataset alignment rules, including variable naming, rescaling and unit normalization, and consistency constraints across datasets. These schemas are then executed within a shared codebase to conduct fusion. This design separates dataset-specific knowledge from the execution layer, ensuring inter-dataset consistency and enabling efficient incorporation of new datasets via schema definitions.

During fusion, three types of datasets are further excluded upon inspection of the standardized representations. First, map-structured datasets with very coarse spatial resolution ($> 5 \text{ km}$, e.g., global-scale products) are excluded, as they may introduce substantial uncertainty due to spatial interpolation. Second, datasets providing long-term past or future projections (typically spanning multiple decades to centuries) are excluded due to their strong dependence on model assumptions and associated uncertainty. Third, sample-structured datasets without precise georeferenced locations are excluded, as they cannot be reliably aligned within the unified sample representation. Finally, \textbf{68 datasets} are integrated into LUCAS-MEGA. These fused datasets are listed in Table~\ref{tab:fused_i} and \ref{tab:fused_ii}.

\subsection{SoilFuser: A Multi-Agent, Human-in-the-Loop Implementation}
\begin{figure}
    \centering
    \includegraphics[width=\linewidth]{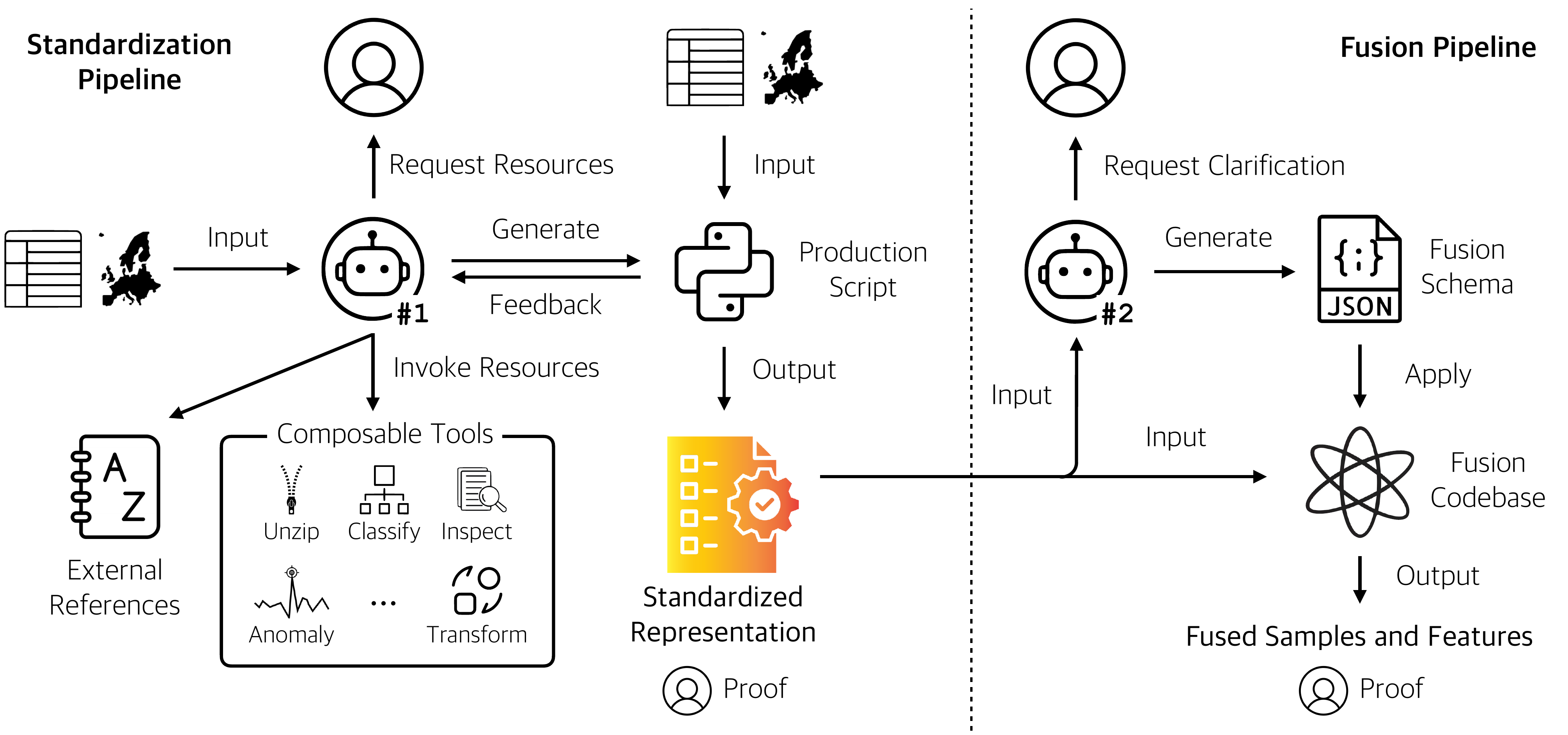}
    \caption{
    Overview of the SoilFuser system, consisting of a standardization pipeline and a fusion pipeline, with AI agents driving their execution. \textbf{Left:} In the standardization pipeline, the agent iteratively generates and executes processing scripts, invoking composable resources and requesting external support when needed to transform heterogeneous datasets into a unified representation. \textbf{Right:} In the fusion pipeline, the agent generates schema mappings that are executed within a shared codebase to integrate the standardized data into a sample--feature representation, forming the final dataset. For both pipelines, the results are examined and validated by human experts.
    }
    \label{fig:fuser}
\end{figure}

While the above procedures define a consistent protocol for data standardization and fusion, their manual execution becomes impractical at the scale and heterogeneity considered in this work. The datasets vary widely in format, structure, and documentation quality, often requiring implicit conventions to be inferred from metadata and external references. We therefore introduce SoilFuser, as outlined in Figure~\ref{fig:fuser}, where AI agents drive the execution of these processes within a human-controlled framework, enabling scalable and consistent integration of heterogeneous data.

In the standardization pipeline, we provide the agent with \textit{resources}, including external references such as codebooks for categorical variables, and a toolbox of composable APIs covering file reading and inspection, data structure recognition, anomaly detection and correction, and file type conversion. Given a clear specification of the target format, the agent orchestrates these resources to generate end-to-end processing scripts for each dataset. Based on execution feedback, it may request additional support from human operators, e.g., to provide missing codebooks or extend available tools, as its ability to create new resources is intentionally restricted to ensure reliability (avoid hallucination). Through this process, heterogeneous data formats, codebooks, and value conventions are normalized into a consistent and inspectable representation.

In the fusion pipeline, the process is similar, except that the agent generates schema definitions instead of processing scripts. Given the standardized representations, the agent specifies how source records are mapped to the unified sample--feature configuration. The generated schemas are then executed to integrate data across datasets. As in standardization, the agent may request external support when schema definitions are ambiguous or incomplete. The agent also attaches natural language annotations to features as part of the fusion process.

\section{The LUCAS-MEGA Dataset}
\label{sec:dataset}

LUCAS-MEGA is organized in an ML-ready tabular representation, where rows correspond to soil samples and columns to features aggregated from multiple sources, while individual cells may contain structured, multimodal data beyond simple scalar values. Cell values span multiple modalities, including numerical variables (scalar and vector-valued), categorical attributes, textual descriptions, and images. 

\subsection{Sample Space}

The sample space of LUCAS-MEGA is primarily populated by the LUCAS topsoil survey, a large-scale, harmonized monitoring program providing georeferenced soil observations across Europe~\citep{orgiazzi2018lucas}, and is further enriched with complementary datasets including EU-HYDI, which provides hydraulic property measurements~\citep{weynants2013euhydi}, and SPADE/M, a European soil profile database with detailed morphological and analytical descriptions~\citep{hiederer2006spadem}. In total, the dataset comprises \textbf{72{,}552 soil samples}, dominated by multiple LUCAS campaigns (21{,}859 from 2015, 21{,}681 from 2009, and 18{,}984 from 2018), along with country-specific extensions from the 2012 campaign (1{,}369 samples from Romania and 661 from Bulgaria) and a Switzerland extension in 2015 (150 samples), complemented by 5{,}580 samples from EU-HYDI and 2{,}268 from SPADE/M.

Each sample in LUCAS-MEGA corresponds to a single soil observation. Multiple samples may share the same geographic location. In the LUCAS survey, this occurs due to repeated visits over time, whereas in EU-HYDI and SPADE/M it arises from measurements at different soil depths. In total, the dataset contains \textbf{29{,}872 unique georeferenced locations}. This design results in a structured long-table representation that avoids internal dimensionalities such as time and depth. Applications that require temporal or depth-aware analysis can be supported through aggregation over samples at the same location.

\subsection{Feature Space}
Features correspond to the columns of our tabular representation, aggregated from heterogeneous data sources (see Tables~\ref{tab:fused_i} and \ref{tab:fused_ii}). In this section, we describe the feature space of LUCAS-MEGA from two complementary perspectives: their semantic meaning as soil and environmental properties, and their ML-related characteristics.

\subsubsection{Properties, Themes and Modalities}

LUCAS-MEGA comprises a total of \textbf{1017 features}, covering a broad range of soil-related characteristics, including intrinsic soil properties (e.g., physical, chemical, biological, and hydrological) and site-specific attributes such as terrain, land cover, and climatic variables. Site-specific attributes are shared across samples collected at the same location.

To facilitate data organization, these features are grouped into \textbf{23 themes}. For example, the feature \textit{clay\_percentage (\%)} is categorized under the \textit{texture} theme, \textit{organic\_carbon\_content (g/kg)} under \textit{carbon}, and \textit{annual\_precipitation (mm)} under \textit{climate}. It should be noted that this classification is intended solely for management and presentation purposes; a given feature may reasonably belong to multiple themes. Therefore, themes do not carry strict semantic meaning but instead serve as an organizational layer. Table~\ref{tab:feature_modalities} shows the feature counts in each theme.

\begin{table}[b!]
\caption{Distribution of feature modalities across themes in LUCAS-MEGA.}
\label{tab:feature_modalities}
\centering
\small
\setlength{\tabcolsep}{4pt}
\renewcommand{\arraystretch}{0.95}
\begin{tabular*}{\textwidth}{@{\extracolsep{\fill}}lcccccc|lcccccc}
\toprule
& Scalar & Vector & Cat. & Text & Image & $\sum$ & & Scalar & Vector & Cat. & Text & Image & $\sum$ \\
\midrule
\textit{ASSETS} & 0 & 3 & 0 & 0 & 1 & 4 & \textit{land\_degradation} & 4 & 0 & 32 & 0 & 0 & 36 \\
\textit{biodiversity} & 15 & 0 & 0 & 0 & 0 & 15 & \textit{land\_site} & 3 & 0 & 17 & 4 & 0 & 24 \\
\textit{biomass} & 20 & 2 & 0 & 0 & 0 & 22 & \textit{management} & 6 & 0 & 7 & 1 & 0 & 14 \\
\textit{carbon} & 43 & 1 & 1 & 0 & 0 & 45 & \textit{mass\_density} & 13 & 0 & 3 & 0 & 0 & 16 \\
\textit{chemical} & 9 & 0 & 0 & 0 & 0 & 9 & \textit{mineral} & 34 & 0 & 3 & 6 & 0 & 43 \\
\textit{climate} & 23 & 3 & 1 & 2 & 0 & 29 & \textit{organic\_matter} & 2 & 0 & 0 & 0 & 0 & 2 \\
\textit{crop\_plant} & 26 & 0 & 0 & 0 & 0 & 26 & \textit{radioactivity} & 4 & 0 & 0 & 0 & 0 & 4 \\
\textit{erosion} & 38 & 14 & 21 & 0 & 0 & 73 & \textit{soil\_type} & 0 & 1 & 13 & 5 & 0 & 19 \\
\textit{fertility} & 45 & 0 & 4 & 0 & 0 & 49 & \textit{texture} & 23 & 0 & 7 & 2 & 0 & 32 \\
\textit{function\_suitability} & 447 & 0 & 5 & 1 & 0 & 453 & \textit{topography\_geology} & 7 & 0 & 14 & 2 & 0 & 23 \\
\textit{greenhouse} & 5 & 0 & 0 & 0 & 0 & 5 & \textit{trace\_elements} & 40 & 0 & 0 & 0 & 0 & 40 \\
\textit{hydraulic} & 13 & 8 & 13 & 0 & 0 & 34 &  &  &  &  &  &  &  \\
\midrule
 &  &  &  &  &  &  & $\sum$ & 820 & 32 & 141 & 23 & 1 & 1017 \\
\bottomrule
\end{tabular*}
\par\vspace{1em}
\noindent\begin{minipage}{\textwidth}
\textit{Note:} ``Scalar'' and ``Vector'' denote scalar- and vector-valued numerical variables, respectively, while ``Cat.'' denotes categorical variables. The \textit{ASSETS} theme manages dense, high-volume data objects: the three vector-valued assets include particle size distributions (LUCAS) and hydraulic conductivity and water retention curves (EU-HYDI), while the image asset consists of site photographs from LUCAS.
\end{minipage}
\end{table}

There are four modalities in LUCAS-MEGA: numerical features (either scalar or vector-valued), categorical features, textual descriptions, and images, with their distribution summarized in Table~\ref{tab:feature_modalities}. Scalar features are single-valued attributes, such as organic carbon content and pH in water, whereas vector-valued features contain multiple entries, such as monthly temperature and precipitation. Categorical features include classifications such as texture class and soil group, textual features include site and sample descriptions, and image features correspond to site photographs. Scalar numerical features dominate the feature space. This multimodal and structurally heterogeneous feature space calls for models that jointly process diverse data types, including numerical encoding, categorical embedding, language understanding, and visual feature extraction.

For semantically associated numerical features, the choice between scalar and vector representations is guided by the structure of the recorded features and their interpretation in modeling. For example, monthly precipitation is represented as a 12-dimensional vector because its monthly values jointly characterize a continuous climatic regime. By contrast, soil texture components, including clay, silt, sand, and coarse contents, are retained as independent scalars---although the texture components are compositionally related, keeping them as separate scalar features allows models to learn their distinct associations with other soil and environmental attributes.

\subsubsection{Missingness and Uncertainty}

\begin{figure}
    \centering
    \includegraphics[width=1.0\linewidth]{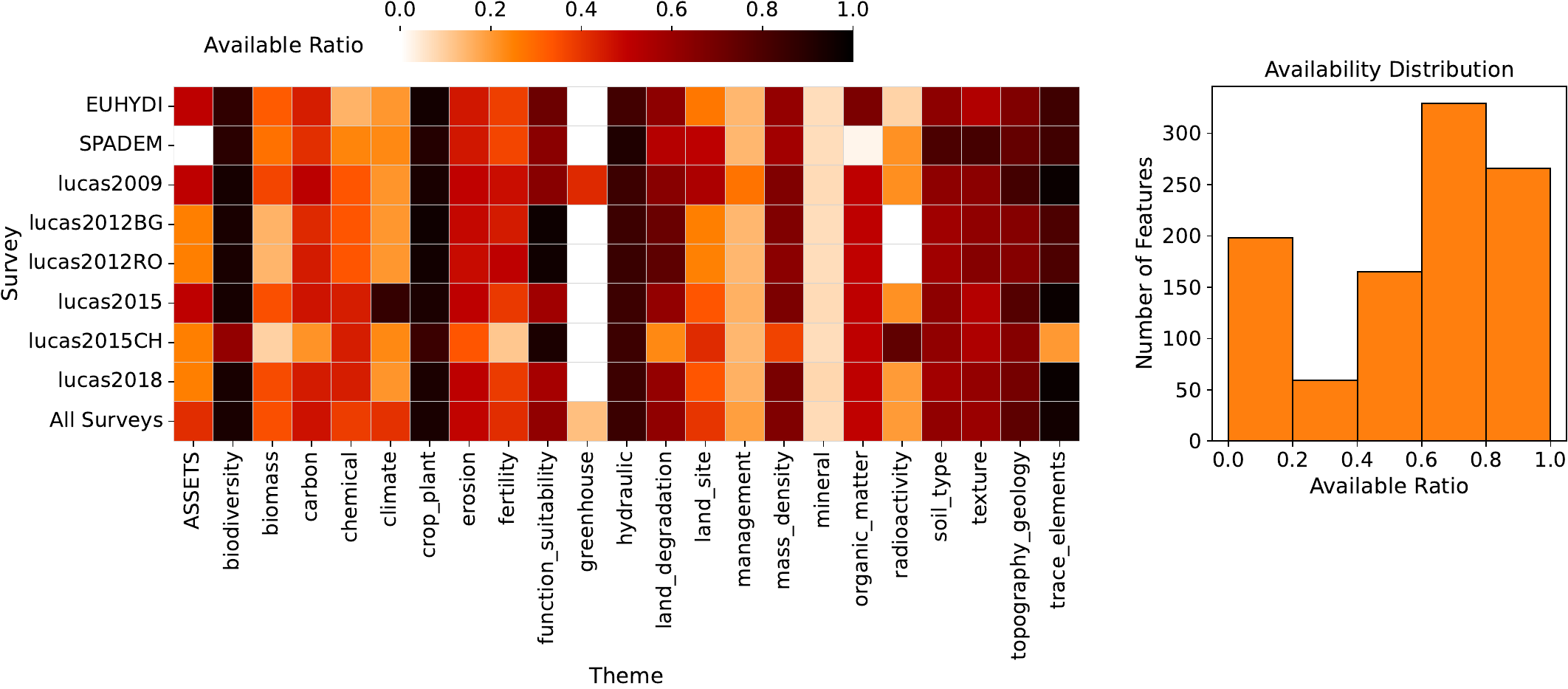}
    \caption{
Missingness of LUCAS-MEGA features. 
\textbf{Left:} Heatmap of feature availability across surveys and thematic groups, where darker shades indicate higher data availability. 
\textbf{Right:} Histogram showing the distribution of availability across all features.
}
    \label{fig:miss}
\end{figure}

Feature missingness refers to the absence of values for certain variables across samples, arising from incomplete measurements or heterogeneous data sources. It is a central challenge for data-driven methods, including machine learning, as it reduces effective sample size, disrupts cross-variable correlations, and introduces bias when missingness is structural. Proper handling of missingness is therefore essential for learning reliable and generalizable representations, particularly in high-dimensional and multimodal settings. Figure~\ref{fig:miss} shows the feature availability of LUCAS-MEGA across surveys and thematic groups. The availability exhibits strong heterogeneity across both features and samples (as reflected by different surveys), with some themes consistently well-covered while others remain sparsely observed. This pattern reflects practical constraints in soil data collection, where field sampling, laboratory measurements, and survey design lead to uneven coverage across variables.

Uncertainty in the source data is sparsely quantified and cannot be systematically recovered. Most datasets lack repeated measurements and do not report measurement uncertainty, limiting direct assessment of observation-level noise. A few model-derived datasets provide uncertainty estimates, which are retained as features. For map-structured data, additional uncertainty arises from spatial alignment during fusion, for which we record the distance between the sample location and the corresponding grid cell. For coarse-resolution maps, this distance can reach several kilometers; note that datasets with excessively low resolution are excluded during fusion. While this distance serves as a proxy for the additional uncertainty introduced during fusion, the intrinsic modeling uncertainty of source data remains largely unquantified.

\subsection{Final Data Formats}

LUCAS-MEGA is released in two complementary formats: a sample-based dictionary stored as JSON and a tabular dataset stored as CSV with accompanying column metadata. The JSON format provides the most complete view of the dataset by organizing samples as dictionary entries, where each feature is stored together with detailed metadata (e.g., units, provenance, and fusion-related attributes), thereby preserving fine-grained information. The CSV format provides a flattened table with samples as rows and features as columns, where metadata are aggregated at the column level for simplicity. For example, source datasets of a feature are consolidated, and spatial alignment distances are summarized using descriptive statistics. This tabular format improves usability and efficiency, at the cost of reduced metadata granularity.

\section{Data-Augmented Reasoning with LUCAS-MEGA}
\label{sec:rag}

Recent advances in large language models (LLMs) have enabled natural language to emerge as a general interface for interacting with complex systems and data. In this context, we develop a unified framework in which natural language serves as an interface for both data access and data-grounded reasoning. Users can directly retrieve soil data through intuitive queries, while the same system supports answering domain-specific questions by grounding responses in retrieved data, improving reliability and interpretability.

To achieve this, LUCAS-MEGA is coupled with structured API layers and integrated into a data-augmented generation framework, in which LLMs act as a natural-language reasoning layer grounded in authoritative data sources. We provide three layers of APIs, all based on reliable data sources:

\begin{enumerate}
    \item \textbf{Geographic reasoning.} This layer resolves user queries into precise spatial references. Natural language place names are first mapped to geographic coordinates via Nominatim~\citep{nominatim}, which are then associated with administrative regions through a hierarchical representation based on GADM~\citep{gadm}. This hierarchy enables consistent spatial grounding and supports reasoning across geographic levels, such as relating areas to their enclosing regions or comparing neighboring administrative units.

    \item \textbf{Feature screening.} Given the high dimensionality of the dataset (over 1,000 features), directly selecting relevant variables from the full feature space is challenging for LLMs due to attention dispersion over long contexts. This layer performs an initial screening step using keyword- and embedding-based matching, mapping user queries to a small set of candidate features with high semantic relevance. By reducing the feature space to a manageable subset, it enables the LLM to make more reliable selections and ensures that subsequent data retrieval remains focused and interpretable.

    \item \textbf{Data retrieval.} Once spatial context and target features are determined, this layer retrieves data from the dataset. Two complementary modes are provided: \textit{a sample-centric mode}, which returns multiple features for a small number of nearby samples, and \textit{a feature-centric mode}, which returns spatial distributions for a small set of features across many locations. This design reflects the practical constraint that queries large in both sample and feature dimensions are difficult to present and interpret in LLM-based interaction. By constraining one dimension, the two modes support common query patterns: detailed characterization at specific locations, or spatial analysis of a small set of variables.
\end{enumerate}

These APIs are exposed via the OpenAPI protocol, enabling integration with LLM-based interfaces. As an example, we deploy ERP-GPT-EU through OpenAI's ChatGPT (see Section~\ref{sec:codedata}).

\section{SoilFormer: A Multimodal Tabular Transformer}
\label{sec:learn}

To demonstrate the utility of LUCAS-MEGA for data-driven modeling, we introduce \textbf{SoilFormer}, a multimodal transformer designed for representation learning on heterogeneous soil--environment data. The inclusion of model training in this work serves two purposes. First, it provides a systematic validation of the dataset by assessing whether meaningful and stable representations can be learned from its high-dimensional, multimodal, and partially observed feature space. Second, it reflects a broader research objective: enabling foundation-model-style learning for soil systems, where complex interactions among diverse variables are captured through large-scale pretraining. While the primary focus of this paper is on dataset construction, SoilFormer establishes a concrete modeling paradigm that leverages the structure and scale of LUCAS-MEGA.

\subsection{Feature Selection}

As the primary contribution of this work is dataset rather than modeling, we adopt a simplified feature selection strategy for pretraining. Specifically, we apply three filtering criteria. First, features with availability below 50\% across samples are excluded to avoid instability caused by highly sparse variables. Second, textual modalities are omitted from the current modeling setup to limit model complexity, as their integration would require combining tabular transformers with large language models. Third, for map-based features, we exclude features whose maximum spatial alignment distance across samples exceeds 200\,m to reduce uncertainty arising from spatial mismatch. After applying these filtering criteria, we obtain the final feature subset used for model training, consisting of 60 numerical features (including both scalar and vector-valued variables), 10 categorical features, and one visual feature, as summarized in Table~\ref{tab:train_feature_1} and \ref{tab:train_feature_2}.

These design choices control modeling complexity by excluding aspects that each require dedicated methodological treatment. In particular, learning under high missing ratios in tabular data often requires dedicated mechanisms that are robust to missingness patterns and missingness shifts~\citep{lee2025mirrams,samad2024ifial}. Likewise, incorporating textual attributes in a principled manner would require coupling tabular modeling with language modeling components~\citep{hegselmann2023tabllm,jaitly2023serialization}. Finally, explicitly handling uncertainty in the inputs would require explicit uncertainty embedding~\citep{valdenegro2025inputuncertainty,buehler2024combining}. These directions are active research topics and would require additional architectural innovations that are outside the main focus of this study. We therefore defer them to future studies to focus on a tractable setting.

\subsection{Architecture}

Transformer-based models~\citep{vaswani2017attention} have become a general-purpose architecture for modeling heterogeneous and multimodal data, owing to their token-based representation and attention mechanism that naturally support variable-length inputs, cross-modal interactions, and partial observations. In particular, their ability to operate on sets of tokens without requiring fixed feature ordering makes them well suited for settings with structured missingness, where subsets of features may be absent or irregularly observed. Building on these advantages, a line of work on tabular transformers~\citep{huang2020tabtransformer,arik2021tabnet,gorishniy2021revisiting} has explored adapting transformer architectures to tabular data, typically by embedding numerical and categorical features as tokens and learning feature-wise interactions through self-attention. These models have demonstrated strong performance and flexibility, especially in capturing complex inter-feature dependencies.

Based on these developments, we design SoilFormer, a multimodal tabular transformer that integrates numerical, categorical, and visual features within a unified token-based framework. Rather than proposing a fundamentally new architecture, we focus on incorporating structural inductive biases tailored to the characteristics of LUCAS-MEGA, while maintaining computational efficiency at scale. Specifically, we introduce the following modifications to improve efficiency and support multimodal inputs, as illustrated in Figure~\ref{fig:arch}:

\begin{figure}
    \centering
    \includegraphics[width=0.9\linewidth]{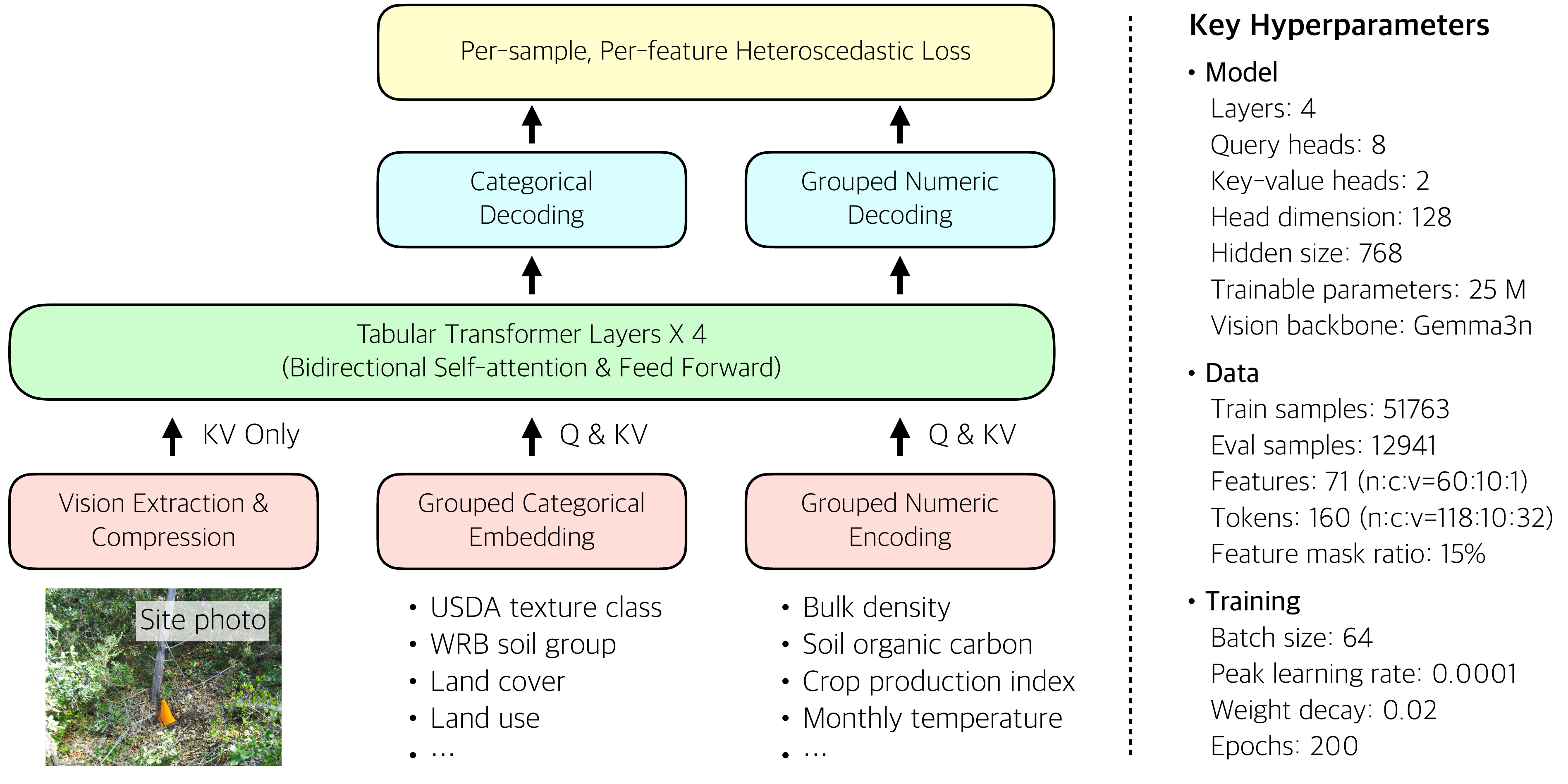}
    \caption{Overview of the SoilFormer architecture. Numerical, categorical, and visual inputs are encoded into a unified token representation via grouped numerical encoding, grouped categorical embedding, and vision feature extraction with latent compression. The tokens are processed by stacked transformer layers, followed by feature-specific decoding heads. Training is driven by masked feature reconstruction, where 15\% of features are randomly masked and reconstructed under a heteroscedastic loss.}
    \label{fig:arch}
\end{figure}

\begin{enumerate}
    \item \textbf{Grouped numerical feature encoding.} In a transformer-based formulation, each feature typically requires its own encoding and decoding heads. We group numerical features by intrinsic dimensionality to enable batched encoding and decoding while preserving feature-wise independence. By avoiding explicit loops over dozens of per-feature heads, this approach improves computational efficiency by an order of magnitude. All numerical features are normalized to z-scores on a per-feature basis.

    \item \textbf{Grouped categorical embedding.} Similarly, categorical features are embedded using a unified embedding table constructed from per-feature vocabularies, enabling efficient lookup while preserving feature-wise parameterization. Decoding logits are computed with separate prediction heads per feature to avoid unnecessary complexity.

    \item \textbf{Vision feature extraction and compression.} Due to the limited scale of available image data, we do not train a vision encoder from scratch. Instead, we use the pretrained vision tower of Gemma3n~\citep{gemma2024} as a frozen feature extractor and treat the extracted visual features as conditioning signals. To align visual and tabular representations while controlling sequence length, we apply a Perceiver-style latent attention module~\citep{jaegle2021perceiver} that compresses 256 visual tokens into 32 latent tokens. This keeps the visual tokens fewer than the tabular tokens (128), avoiding visual-modality dominance while preserving cross-modal interaction.
\end{enumerate}

\subsection{Masked Feature Modeling with Heteroscedastic Loss}

We adopt masked feature modeling (MFM) as the self-supervised training objective, analogous to masked language modeling (MLM) in BERT~\citep{devlin2019bert}. Samples in LUCAS-MEGA already contain native missingness due to incomplete observations across heterogeneous sources. During training, we further mask 15\% of the observed features at random and train the model to reconstruct them from the remaining context. Consequently, each training input contains two forms of absent features: native missing features and actively masked features. Both are presented as missing values at the input level, but supervision is applied only to the actively masked subset, for which ground-truth values are available. This formulation enables the model to learn cross-feature dependencies from partially observed multimodal inputs.

A key design choice is that the model does not distinguish between native missingness and active masking at the encoding stage. For both numerical and categorical variables, all missing entries are represented using dedicated per-feature missing embeddings, yielding a unified representation of missingness. The distinction between the two types of missingness is retained only in the loss: reconstruction is supervised on actively masked features. This choice aligns training with inference: at inference time, no artificial masking is applied, and native missing entries are encoded using the same per-feature missing embeddings as in training.

To account for the varying noise levels across features and samples in LUCAS-MEGA, we further adopt a heteroscedastic objective~\citep{kendall2017uncertainties}. In contrast to standard reconstruction losses that weight all targets equally, this formulation uses predicted uncertainty to adaptively weight reconstruction errors, down-weighting noisy or intrinsically difficult predictions. Specifically, SoilFormer predicts not only the reconstructed value or class logits, but also a scalar $s$ for each masked feature in each sample. This scalar represents the per-feature, per-sample log-variance of the prediction, with the predictive variance given by $\sigma^2 = \exp(s)$. This learned uncertainty serves two roles. During training, it stabilizes optimization by balancing contributions from masked targets with different noise levels and reconstruction difficulty. After training, it provides an uncertainty-aware characterization of the representation, indicating where the model considers reconstruction to be confident or ambiguous.

The heteroscedastic objective is instantiated separately for numerical and categorical features. For numerical features, the decoder predicts a mean $\mu$ and a log-variance $s$, and the loss is given by the Gaussian negative log-likelihood
\begin{equation}
\mathcal{L}_{\mathrm{num}}
=
(y-\mu)^2/\exp(s) + s,
\label{eq:num}
\end{equation}
where $y$ is the ground-truth target. The factor $\exp(s)$ adaptively rescales the reconstruction error, while the additive $s$ term prevents the model from trivially inflating uncertainty. For categorical features, we use an analogous uncertainty-weighted form of cross-entropy,
\begin{equation}
\mathcal{L}_{\mathrm{cat}}
=
\mathrm{CE}(p,y)/\exp(s) + s,
\label{eq:lat}
\end{equation}
where $p$ denotes the predicted class distribution. This formulation follows the heuristic approximation in~\cite{kendall2017uncertainties}, which improves optimization stability compared to using $\exp(s)$ as a temperature parameter in the cross-entropy loss. The final objective sums $\mathcal{L}_{\mathrm{num}}$ and $\mathcal{L}_{\mathrm{cat}}$ over all actively masked features in each sample.

\subsection{Model Evaluation}

\subsubsection{Pretraining Behavior}

\begin{figure}
    \centering
    \includegraphics[width=\linewidth]{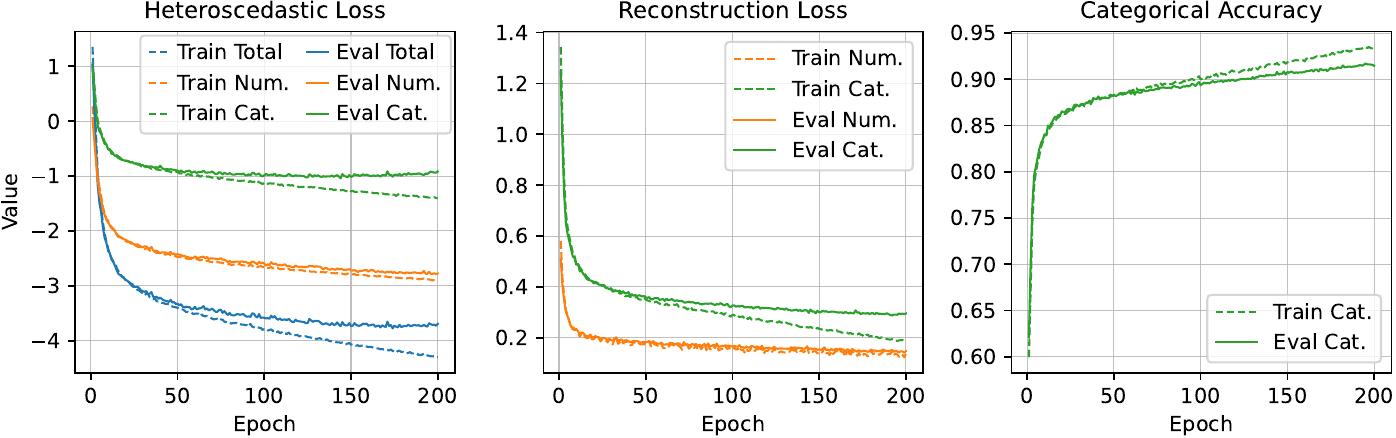}
    \caption{
    Convergence of losses and accuracy during pretraining. \textbf{Left:} Heteroscedastic losses correspond to the actual training objective, as defined in Eqs.~\eqref{eq:num} and \eqref{eq:lat}. \textbf{Middle:} Reconstruction losses denote the underlying mean squared error (MSE) and cross-entropy (CE) terms in these equations, i.e., the reconstruction error before uncertainty reweighting. \textbf{Right:} Categorical accuracy measures the top-1 accuracy over masked categorical features.
}
    \label{fig:loss}
\end{figure}

Figure~\ref{fig:loss} shows stable pretraining dynamics over 200 epochs. For numerical features, both the heteroscedastic loss and the underlying reconstruction loss decrease smoothly on the training and evaluation sets, with a modest train-evaluation gap emerging at later epochs. This behavior indicates harmonized improvement in reconstruction accuracy and predictive confidence, suggesting that the numerical branch remains well calibrated under the heteroscedastic formulation throughout training.

The categorical branch exhibits a different but still interpretable behavior. Its training loss continues to decrease, while the evaluation loss begins to rise slightly in later epochs, even as evaluation accuracy continues to improve. This combination indicates that categorical predictions become increasingly accurate in terms of top-1 classification, but less well calibrated probabilistically. In other words, the model increasingly identifies the correct class, yet assigns less concentrated probability mass to a subset of harder or noisier examples, causing cross-entropy to increase despite improving accuracy. This is precisely where the heteroscedastic formulation becomes useful: by explicitly modeling prediction uncertainty, it allows training to remain stable and exposes the trade-off between accuracy and confidence at the checkpoint level. We therefore select 200 epochs as a practical operating point, where the model achieves high categorical accuracy while avoiding the stronger calibration drift observed in longer runs.

\subsubsection{Predictive Performance}

\begin{figure}
    \centering
    \includegraphics[width=\linewidth]{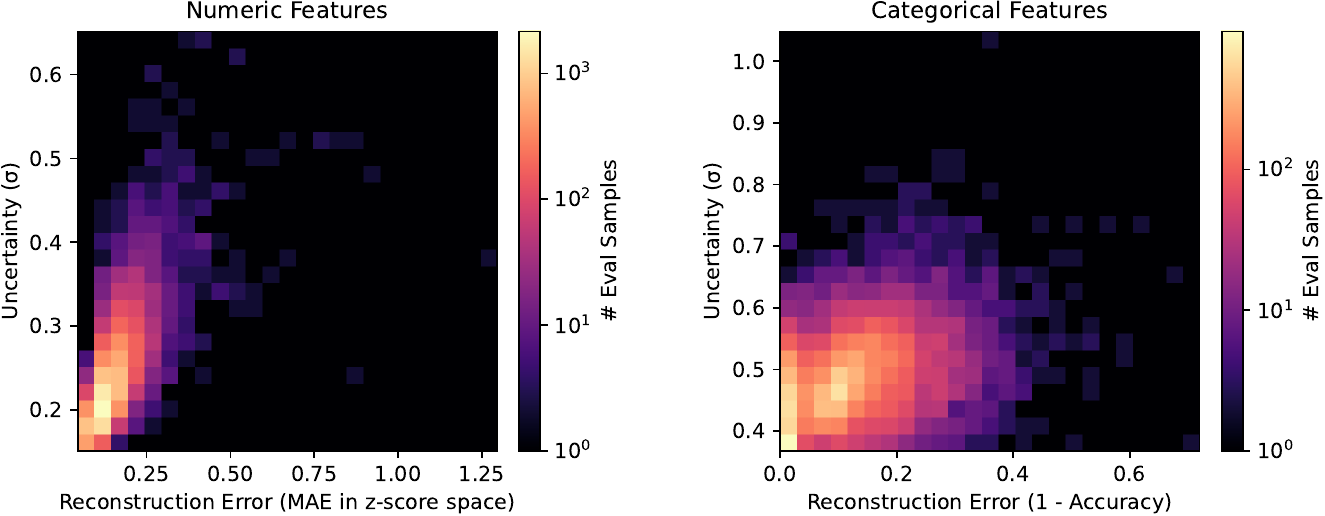}
    \caption{
Distributions of evaluation samples in the error--uncertainty landscape.
For each sample in the evaluation set, reconstruction errors and predictive uncertainties are aggregated over masked features across 100 random masking trials, each with a masking ratio of 15\%.
\textbf{Left:} For numerical features, reconstruction error (x-axis) is measured as mean absolute error (MAE) in the z-score space.
\textbf{Right:} For categorical features, reconstruction error (x-axis) is measured as classification error ($1-\mathrm{accuracy}$).
In both cases, uncertainty (y-axis) is given by the learned heteroscedastic standard deviation $\sigma = \exp(s/2)$.
The 2D histograms show the number of evaluation samples falling into each error--uncertainty bin, with color indicating sample counts on a logarithmic scale.
}
\label{fig:sample}
\end{figure}

Figure~\ref{fig:sample} presents the distribution of evaluation samples in the error--uncertainty landscape. To construct this figure, we perform 100 random masking trials on the evaluation set, each with a masking ratio of 15\%. In each trial, the model predicts the masked features and their associated heteroscedastic uncertainties. The resulting reconstruction errors and uncertainties are aggregated at the sample level across masked features and trials. Samples are then grouped into bins in the error--uncertainty space, and the number of samples in each bin is visualized as a 2D histogram. In this representation, the horizontal axis reflects predictive accuracy, whereas the vertical axis reflects predictive uncertainty.

A dominant concentration of samples is observed in the low-error, low-uncertainty region (i.e., the lower-left corner). This pattern suggests that a large portion of the dataset admits accurate and confident reconstruction under partial observation, indicating strong statistical dependencies among features. From a data perspective, this supports the presence of a coherent underlying structure in soil--environment systems, where variables exhibit strong interdependencies. From a modeling perspective, it provides empirical evidence that the proposed architecture and heteroscedastic objective are well aligned with the data characteristics, enabling the model to learn representations that are both predictive and uncertainty-aware.

\subsubsection{Feature Interaction Analysis}

\begin{figure}
    \centering
    \includegraphics[width=\linewidth]{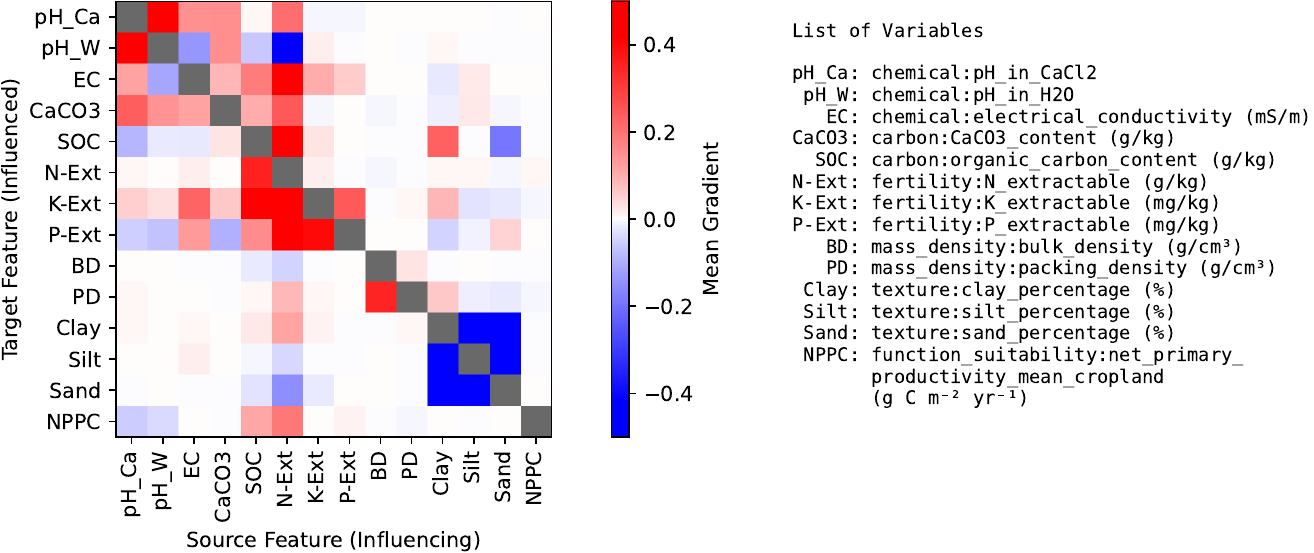}
\caption{
Mask-conditioned Jacobian (MCJ) matrix estimated from SoilFormer. For each target feature, its value is individually masked and predicted by the model from the remaining inputs on the evaluation set. The gradient of the predicted target with respect to each source feature is then computed by automatic differentiation in the z-score-normalized space and averaged across samples. Repeating this procedure over all considered target features yields the MCJ matrix, which summarizes sample-averaged directional sensitivities and is generally asymmetric. Rows denote target features, columns denote source features, and the feature name mapping is listed on the right.
}
\label{fig:grad}
\end{figure}

Beyond overall predictive performance, Figure~\ref{fig:grad} is designed to probe how SoilFormer captures the interdependencies among soil properties. We examine the directional sensitivity of each predicted feature to the available inputs. The resulting mask-conditioned Jacobian (MCJ) matrix provides a compact view of how strongly one variable influences another across the evaluation set, serving as a first lens into the feature interaction structure learned by the model.

Several patterns in Figure~\ref{fig:grad} are consistent with established knowledge and provide a first validation that the model has learned meaningful structure:
\begin{enumerate}
    \item Several near-trivial relationships are correctly recovered. \texttt{pH\_Ca} and \texttt{pH\_W} show strong positive mutual associations, \texttt{BD} and \texttt{PD} are strongly positively related, and \texttt{Clay}, \texttt{Silt}, and \texttt{Sand} exhibit strong mutual trade-offs. These basic relationships serve as an immediate sanity check that the model has captured obvious data regularities.
    
    \item The interaction patterns between \texttt{SOC} and soil texture are consistent with established soil knowledge. In particular, \texttt{SOC} increases strongly with \texttt{Clay}, increases more weakly with \texttt{Silt}, and decreases with \texttt{Sand}, consistent with the stronger carbon stabilization capacity of finer-textured soils.
    
    \item The cropland productivity proxy \texttt{NPPC} shows strong positive dependence on \texttt{SOC} and \texttt{N-Ext}, weaker positive dependence on \texttt{K-Ext} and \texttt{P-Ext}, and a negative association with pH. This suggests that the model plausibly links soil carbon and nutrient status to productive ecosystem function.
    
    \item \texttt{N-Ext} appears as a broad influencer across many target variables, suggesting that the model assigns concentrated predictive attention to nitrogen-related information. In particular, the strong coupling between \texttt{N-Ext} and \texttt{SOC} is notable and is consistent with the close linkage between soil organic matter status and nitrogen availability under management.
\end{enumerate}

This analysis should nevertheless be regarded as a preliminary exploration of model behavior. First, the MCJ matrix is based on sample-averaged local gradients evaluated at many different points in feature space, and therefore reflects only coarse global tendencies rather than the response at any individual sample. Second, these quantities represent local predictive sensitivities rather than causal effects. A causal analysis would require controlled interventions to test how the presence or absence of a variable changes both the predicted value and the associated uncertainty. More generally, relating learned model behavior to objective soil mechanisms would require designs tailored to specific variable relationships, including targeted sample selection (e.g., focusing on subsets where a variable spans informative ranges) and controlled masking schemes. Such investigations are problem-specific and are left for future modeling studies.

\section{Conclusions}

We introduced LUCAS-MEGA, a large-scale multimodal dataset for soil--environment systems constructed through systematic fusion of heterogeneous European soil--environment observations, with the LUCAS survey as its backbone. By integrating 68 datasets from ESDAC into a unified sample--feature representation, LUCAS-MEGA comprises over 70,000 samples and more than 1,000 features spanning numerical, categorical, textual, and visual modalities. To enable this construction at scale, we developed SoilFuser, a multi-agent, human-in-the-loop pipeline for data standardization and fusion. We further demonstrated the utility of LUCAS-MEGA for multimodal representation learning through SoilFormer, showing that stable self-supervised pretraining is feasible on this heterogeneous, partially observed feature space, and that the resulting representations capture meaningful soil--environment relationships. Together, these results establish LUCAS-MEGA as both a practical data resource for soil science and a useful testbed for multimodal learning under real-world missingness and uncertainty.

Several directions follow from this work. First, the standardized and schema-driven design of SoilFuser makes LUCAS-MEGA readily extensible as new soil--environment data sources become available. Within Europe, for example, this includes the LUCAS 2022 topsoil survey, planned for release in Q4 2026. More broadly, the same framework could support expansion toward a global soil dataset by integrating regional and international data sources. Second, the present modeling study adopts a simplified setting to focus on the dataset contribution. Important next steps therefore include extending SoilFormer to handle the components that were excluded or simplified here, particularly textual modalities, features with very low availability, and variables associated with high input uncertainty. Addressing these aspects will require more advanced architectures for language--tabular integration, robust learning under extreme missingness, and explicit treatment of input uncertainty. These challenges provide a natural agenda for future research built upon LUCAS-MEGA.

\section{Code and Data Availability}
\label{sec:codedata}

\begin{itemize}

\item \textbf{LUCAS-MEGA} (\url{https://huggingface.co/datasets/earthroverprogram/lucas-mega}) \quad 
Integrated sample--feature representations in tabular and dictionary formats, with unified metadata and asset files. Additional construction materials include intermediate standardization outputs, together with the scripts and schema definitions used to construct LUCAS-MEGA.

\item \textbf{SoilFormer} (\url{https://huggingface.co/earthroverprogram/soilformer}) \quad
SoilFormer architecture, pretrained weights, and scripts for training, evaluation, and inference.

\item \textbf{ERP-GPT-EU} (\url{https://huggingface.co/datasets/earthroverprogram/erp-gpt-eu}) \quad
Data-query APIs, prompts, knowledge files, and related resources for ERP-GPT-EU.

\end{itemize}

\appendix
\section{Source Datasets of LUCAS-MEGA}
Table~\ref{tab:fused_i} and \ref{tab:fused_ii} contain all source datasets integrated into LUCAS-MEGA.

\begin{table}[htbp]
\caption{Source datasets integrated into LUCAS-MEGA (Part I).}
\label{tab:fused_i}
\centering
\renewcommand{\arraystretch}{0.9}
\setlength{\tabcolsep}{0pt}
\begin{tabular*}{\textwidth}{@{\extracolsep{\fill}\hspace{1em}}l@{}}
\toprule
Dataset ID \\
\midrule
3d-soil-hydraulic-database-europe-1-km-and-250-m-resolution \\
arsenic-european-topsoils \\
bacterial-fungal-biomass-fatty-acid-methyl-esters \\
biodiversity-factor-soil-erosion \\
cadmium-topsoils-european-union \\
caesium-137-and-plutonium-239240-european-topsoils \\
carbon-budget-eu-agricultural-soils \\
clay-mineral-inventory-soils-europe-based-lucas-2015-survey-soil-samples-2 \\
copper-distribution-topsoils \\
cover-crops-accross-europe \\
cover-management-factor-c-factor-eu \\
estimate-net-erosion-and-sediment-transport-using-watemsedem-european-union \\
european-food-safety-authority-efsa-data-persam-software-tool \\
european-hydropedological-data-inventory-eu-hydi-database-0 \\
european-map-soil-suitability-provide-platform-most-human-activities-eu28 \\
european-soil-database-derived-data \\
european-soil-database-v2-raster-library-1kmx1km \\
global-rainfall-erosivity \\
global-soil-erodibility \\
global-soil-organic-carbon-estimates \\
glosem \\
gully-erosion-based-lucas \\
heavy-metals-topsoils \\
land-degradation-europe \\
land-suitability-temperate-europe \\
ls-factor-slope-length-and-steepness-factor-eu \\
lucas-2009-topsoil-data \\
lucas-2015-topsoil-data-switzerland \\
lucas-2018-topsoil-data \\
lucas2015-topsoil-data \\
map-indicating-availability-raw-material-soils-european-union-organic-soil-material-b-soil \\
maps-heavy-metals-soils-eu-based-lucas-2009-hm-data-0 \\
maps-indicators-soil-hydraulic-properties-europe \\
maps-related-predicting-preservation-cultural-artefacts-and-buried-materials-soils-eu-0 \\
\bottomrule
\end{tabular*}
\end{table}

\begin{table}[htbp]
\caption{Source datasets integrated into LUCAS-MEGA (Part II).}
\label{tab:fused_ii}
\centering
\renewcommand{\arraystretch}{0.9}
\setlength{\tabcolsep}{0pt}
\begin{tabular*}{\textwidth}{@{\extracolsep{\fill}\hspace{1em}}l@{}}
\toprule
Dataset ID \\
\midrule
maps-storing-and-filtering-capacity-soils-europe \\
mercury-content-european-union-topsoil \\
multiple-concurrent-soil-erosion-processes \\
n2o-emissions-agricultural-soils-europe \\
natural-susceptibility-soil-compaction-europe \\
observedtypical-soc-index \\
octop-topsoil-organic-carbon-content-europe \\
pan-european-soc-stock-agricultural-soils \\
pan-european-soil-erosion-risk-assessment-pesera \\
phosphorus-budget-and-p-stocks \\
phosphorus-cycle-european-agricultural-soils \\
potential-threats-soil-biodiversity-europe \\
priming-effects-soils-across-europe \\
rainfall-erosivity-european-union-and-switzerland \\
SOC-changes-2009-18 \\
soil-biomass-productivity-maps-grasslands-and-pasture-coplands-and-forest-areas-european \\
soil-bulk-density-europe \\
soil-carbon-risk-index \\
soil-degradation-indicators-eu \\
soil-erodibility-k-factor-high-resolution-dataset-europe \\
soil-erosion-forestland-europe-using-rusle2015 \\
soil-erosion-water-rusle2015 \\
soil-function-data-measured-lucas-2018-sites-across-eu \\
soil-ghg-fluxes-using-lucas-soil-daycent \\
soil-loss-due-crop-harvesting-european-union \\
soil-microbial-biomass-and-respiration \\
soil-organic-carbon-saturation-capacity \\
soil-organic-matter-som-fractions \\
Soil\_erosion\_by\_wind \\
spadem \\
support-practices-factor-p-factor-eu \\
topsoil-soil-organic-carbon-lucas-eu25 \\
un-sustainable-development-goal-1531-assessment-land-degradation-indicator-eu-scale \\
zn-concentrations-eu-topsoils \\
\bottomrule
\end{tabular*}
\par\vspace{1em}
\noindent\begin{minipage}{\textwidth}
\textit{Note:} Dataset details can be accessed by appending the dataset ID to the URL \url{https://esdac.jrc.ec.europa.eu/content/}.
\end{minipage}
\end{table}

\section{Features for Training SoilFormer}
Table~\ref{tab:train_feature_1} and \ref{tab:train_feature_1} show all features used for training SoilFormer.


\begin{table}[htbp]
\caption{Feature subset used for training SoilFormer (Part I).}
\label{tab:train_feature_1}
\centering
\small
\renewcommand{\arraystretch}{0.9}
\setlength{\tabcolsep}{4pt}
\begin{tabular*}{\textwidth}{@{\extracolsep{\fill}}llll}
\toprule
\textbf{Theme} & \textbf{Feature} & \textbf{Modality} & \textbf{\# Tokens} \\
\midrule
carbon & CaCO3\_content (g/kg) & Numeric (scalar) & 1 \\
 & SOC\_saturation\_ratio & Numeric (scalar) & 1 \\
 & organic\_carbon\_content (g/kg) & Numeric (scalar) & 1 \\
chemical & electrical\_conductivity (mS/m) & Numeric (scalar) & 1 \\
 & pH\_in\_CaCl2 & Numeric (scalar) & 1 \\
 & pH\_in\_H2O & Numeric (scalar) & 1 \\
climate & monthly\_precipitation\_JAN\_to\_DEC (mm) & Numeric (vector) & 12 \\
 & monthly\_temperature\_JAN\_to\_DEC (°C) & Numeric (vector) & 12 \\
crop\_plant & cover\_crop\_fraction\_median (‱) & Numeric (scalar) & 1 \\
 & cover\_crop\_fraction\_std (‱) & Numeric (scalar) & 1 \\
erosion & c\_factor\_cover\_management & Numeric (scalar) & 1 \\
 & ls\_factor\_slope\_length\_and\_steepness & Numeric (scalar) & 1 \\
 & net\_erosion\_rate\_WaTEM\_SEDEM (t ha${}^{-1}$ yr${}^{-1}$) & Numeric (scalar) & 1 \\
 & p\_factor\_support\_practices & Numeric (scalar) & 1 \\
 & soil\_displacement\_estimate (t ha${}^{-1}$ yr${}^{-1}$) & Numeric (scalar) & 1 \\
 & soil\_loss\_by\_water (t ha${}^{-1}$ yr${}^{-1}$) & Numeric (scalar) & 1 \\
fertility & K\_extractable (mg/kg) & Numeric (scalar) & 1 \\
 & N\_extractable (g/kg) & Numeric (scalar) & 1 \\
 & P\_available\_stock (kg ha${}^{-1}$) & Numeric (scalar) & 1 \\
 & P\_extractable (mg/kg) & Numeric (scalar) & 1 \\
 & P\_total\_concentration\_topsoil (mg/kg) & Numeric (scalar) & 1 \\
function\_suitability & cultural\_material\_preservation\_capacity\_index\_bones (0-3) & Numeric (scalar) & 1 \\
 & cultural\_material\_preservation\_capacity\_index\_metals (0-3) & Numeric (scalar) & 1 \\
 & cultural\_material\_preservation\_capacity\_index\_organics (0-3) & Numeric (scalar) & 1 \\
 & cultural\_material\_preservation\_capacity\_index\_stratigraphic\_evidence (0-3) & Numeric (scalar) & 1 \\
 & filter\_capacity\_index\_anion (0-10) & Numeric (scalar) & 1 \\
 & filter\_capacity\_index\_cation (0-10) & Numeric (scalar) & 1 \\
 & filter\_capacity\_index\_non\_aqueous\_phase\_liquids (0-10) & Numeric (scalar) & 1 \\
 & filter\_capacity\_index\_non\_polar\_organics (0-10) & Numeric (scalar) & 1 \\
 & filter\_capacity\_index\_solids\_pathogens (0-10) & Numeric (scalar) & 1 \\
 & net\_primary\_productivity\_mean\_cropland (g C m${}^{-2}$ yr${}^{-1}$) & Numeric (scalar) & 1 \\
 & raw\_material\_availability\_index\_construction\_material (0-10) & Numeric (scalar) & 1 \\
 & raw\_material\_availability\_index\_organics (0-10) & Numeric (scalar) & 1 \\
 & storing\_capacity\_index\_anion (0-10) & Numeric (scalar) & 1 \\
 & storing\_capacity\_index\_cation (0-10) & Numeric (scalar) & 1 \\
\bottomrule
\end{tabular*}
\end{table}

\begin{table}[htbp]
\caption{Feature subset used for training SoilFormer (Part II).}
\label{tab:train_feature_2}
\centering
\small
\renewcommand{\arraystretch}{0.9}
\setlength{\tabcolsep}{4pt}
\begin{tabular*}{\textwidth}{@{\extracolsep{\fill}}llll}
\toprule
\textbf{Theme} & \textbf{Feature} & \textbf{Modality} & \textbf{\# Tokens} \\
\midrule
function\_suitability & storing\_capacity\_index\_non\_aqueous\_phase\_liquids (0-10) & Numeric (scalar) & 1 \\
& storing\_capacity\_index\_non\_polar\_organics (0-10) & Numeric (scalar) & 1 \\
 & storing\_capacity\_index\_solids\_pathogens (0-10) & Numeric (scalar) & 1 \\
 & dominant\_limitation\_to\_agricultural\_use & Categorical & 1 \\
 & secondary\_limitation\_to\_agricultural\_use & Categorical & 1 \\
 & soil\_structure\_for\_agricultural\_use\_subsoil & Categorical & 1 \\
 & soil\_structure\_for\_agricultural\_use\_topsoil & Categorical & 1 \\
 & soil\_suitability\_for\_human\_activities & Categorical & 1 \\
geographic & latitude (deg) & Numeric (scalar) & 1 \\
 & longitude (deg) & Numeric (scalar) & 1 \\
hydraulic & available\_water\_capacity\_profile (cm${}^3$ cm${}^{-3}$ \%) & Numeric (vector) & 7 \\
 & field\_capacity\_water\_content\_profile (cm${}^3$ cm${}^{-3}$ \%) & Numeric (vector) & 7 \\
 & hydraulic\_conductivity\_at\_field\_capacity\_profile (cm/day) & Numeric (vector) & 7 \\
 & saturated\_hydraulic\_conductivity\_profile (cm/day) & Numeric (vector) & 7 \\
 & saturated\_water\_content\_profile (cm${}^3$ cm${}^{-3}$ \%) & Numeric (vector) & 7 \\
 & wilting\_point\_water\_content\_profile (cm${}^3$ cm${}^{-3}$ \%) & Numeric (vector) & 7 \\
land\_site & land\_cover\_primary & Categorical & 1 \\
 & land\_use\_primary & Categorical & 1 \\
mass\_density & bulk\_density (g/cm${}^3$) & Numeric (scalar) & 1 \\
 & bulk\_density\_0\_10cm (g/cm${}^3$) & Numeric (scalar) & 1 \\
 & bulk\_density\_10\_20cm (g/cm${}^3$) & Numeric (scalar) & 1 \\
 & packing\_density (g/cm${}^3$) & Numeric (scalar) & 1 \\
soil\_type & WRB\_soil\_group & Categorical & 1 \\
texture & clay\_percentage (\%) & Numeric (scalar) & 1 \\
 & coarse\_percentage (\%) & Numeric (scalar) & 1 \\
 & sand\_percentage (\%) & Numeric (scalar) & 1 \\
 & silt\_percentage (\%) & Numeric (scalar) & 1 \\
 & ISSS\_class & Categorical & 1 \\
 & USDA\_class & Categorical & 1 \\
topography\_geology & elevation (m) & Numeric (scalar) & 1 \\
 & slope (deg) & Numeric (scalar) & 1 \\
trace\_elements & As\_concentration\_mean (log10 mg/kg) & Numeric (scalar) & 1 \\
 & Cd\_concentration (mg/kg) & Numeric (scalar) & 1 \\
 & Hg\_concentration (µg/kg) & Numeric (scalar) & 1 \\
 & Zn\_concentration (mg/kg) & Numeric (scalar) & 1 \\
ASSETS & sample\_site\_photo & Visual & 32 \\
\midrule
Total & 71 & 4 & 160 \\
\bottomrule
\end{tabular*}
\par\vspace{1em}
\noindent\begin{minipage}{\textwidth}
\end{minipage}
\end{table}


\noappendix       





\clearpage

\authorcontribution{
KL constructed the dataset and conducted the transformer-based representation-learning experiments.
SJ contributed domain expertise in soil science and data interpretation, and provided guidance on data selection, processing choices, model validation and interpretation.
PP contributed through the long-term collection, curation and stewardship of ESDAC source datasets, and provided expertise on European soil data resources and their interpretation.
TN provided senior scientific leadership for the overall project, helped define the data and modeling objectives, and contributed to the design of the representation-learning framework.
KL wrote the initial manuscript draft. All authors contributed to manuscript review, editing and discussion of the results.
}

\competinginterests{We declare that no competing interests are present.}

\disclaimer{LUCAS-MEGA is constructed from publicly available datasets provided by the European Soil Data Centre (ESDAC) and associated data contributors. The original data remain the responsibility of the respective providers. The authors assume no responsibility for errors or inconsistencies in the source data.}

\acknowledgements{
The Earth Rover Program is a not-for-profit organization funded by the Bezos Earth Fund, Founders Pledge, and UBS Optimus Foundation. The authors gratefully acknowledge the European Soil Data Centre (ESDAC) and all contributing data providers for making a broad collection of European soil and environmental datasets publicly available. KL thanks Jiaoyao Meng and Tina Fallah for their assistance with requesting and downloading the source datasets.
}




\bibliographystyle{copernicus}
\bibliography{ref}

@article{turner2021complexity,
  author    = {Turner, Benjamin L.},
  title     = {Soil as an Archetype of Complexity: A Systems Approach to Improve Insights, Learning, and Management of Coupled Biogeochemical Processes and Environmental Externalities},
  journal   = {Soil Systems},
  year      = {2021},
  volume    = {5},
  number    = {3},
  pages     = {39},
  doi       = {10.3390/soilsystems5030039}
}

@article{young2004selforganization,
  author    = {Young, I. M. and Crawford, J. W.},
  title     = {Interactions and Self-Organization in the Soil-Microbe Complex},
  journal   = {Science},
  year      = {2004},
  volume    = {304},
  number    = {5677},
  pages     = {1634--1637},
  doi       = {10.1126/science.1097394}
}

@book{jenny1941factors,
  author    = {Jenny, Hans},
  title     = {Factors of Soil Formation: A System of Quantitative Pedology},
  publisher = {McGraw-Hill},
  address   = {New York},
  year      = {1941}
}

@inproceedings{vaswani2017attention,
  author    = {Vaswani, Ashish and Shazeer, Noam and Parmar, Niki and Uszkoreit, Jakob and Jones, Llion and Gomez, Aidan N. and Kaiser, {\L}ukasz and Polosukhin, Illia},
  title     = {Attention Is All You Need},
  booktitle = {Advances in Neural Information Processing Systems},
  volume    = {30},
  pages     = {5998--6008},
  year      = {2017}
}

@article{bommasani2021opportunities,
  author    = {Bommasani, Rishi and Hudson, Drew A. and Adeli, Ehsan and Altman, Russ and Arora, Simran and von Arx, Sydney and Bernstein, Michael S. and Bohg, Jeannette and Bosselut, Antoine and Brunskill, Emma and Brynjolfsson, Erik and et al.},
  title     = {On the Opportunities and Risks of Foundation Models},
  journal   = {arXiv preprint arXiv:2108.07258},
  year      = {2021}
}

@article{lam2023graphcast,
  author    = {Lam, Remi and Sanchez-Gonzalez, Alvaro and Willson, Matthew and Wirnsberger, Peter and Fortunato, Meire and Alet, Ferran and Ravuri, Suman and Ewalds, Timo and Eaton-Rosen, Zach and Hu, Weihua and Keenan, Timothy and Clancy, Ellen and Mohan, Ankur and Clark, Sam and Deasy, Daniel and Vinyals, Oriol and Heess, Nicolas and Battaglia, Peter and Hassabis, Demis and et al.},
  title     = {Learning skillful medium-range global weather forecasting},
  journal   = {Science},
  volume    = {382},
  number    = {6677},
  pages     = {1416--1421},
  year      = {2023},
  doi       = {10.1126/science.adi2336}
}

@article{poggio2021soilgrids2,
  title   = {SoilGrids 2.0: producing soil information for the globe with quantified spatial uncertainty},
  author  = {Poggio, L. and de Sousa, L. M. and Batjes, N. H. and Heuvelink, G. B. M. and Kempen, B. and Ribeiro, E. and Rossiter, D.},
  journal = {SOIL},
  volume  = {7},
  pages   = {217--240},
  year    = {2021},
  doi     = {10.5194/soil-7-217-2021}
}

@misc{faoiiasa2012hwsd,
  author = {{FAO} and {IIASA}},
  title  = {Harmonized World Soil Database (HWSD)},
  year   = {2012},
  note   = {Version 1.2}
}

@article{king1994esdb,
  title   = {Development of a soil geographic database from the Soil Map of the European Communities},
  author  = {King, D. and Daroussin, J. and Tavernier, R.},
  journal = {Catena},
  volume  = {21},
  number  = {1},
  pages   = {37--56},
  year    = {1994},
  doi     = {10.1016/0341-8162(94)90030-2}
}

@article{batjes2017wosis,
  title   = {WoSIS: providing standardised soil profile data for the world},
  author  = {Batjes, N. H. and Ribeiro, E. and van Oostrum, A. and Leenaars, J. G. B. and Hengl, T. and Mendes de Jesus, J. S.},
  journal = {Earth System Science Data},
  volume  = {9},
  pages   = {1--14},
  year    = {2017},
  doi     = {10.5194/essd-9-1-2017}
}

@article{hiederer2006spadem,
  title   = {Soil Profile Analytical Database for Europe (SPADE): Reconstruction and Validation of the Measured Data (SPADE/M)},
  author  = {Hiederer, R. and Jones, R. J. A. and Daroussin, J.},
  journal = {Geografisk Tidsskrift},
  volume  = {106},
  number  = {1},
  pages   = {71--85},
  year    = {2006},
  doi     = {10.1080/00167223.2006.10649546}
}

@techreport{weynants2013euhydi,
  title       = {European HYdropedological Data Inventory (EU-HYDI)},
  author      = {Weynants, M. and Montanarella, L. and T{\'o}th, G.},
  institution = {Publications Office of the European Union},
  year        = {2013},
  doi         = {10.2788/5936}
}

@article{hengl2015afsis,
  title   = {Mapping Soil Properties of Africa at 250 m Resolution},
  author  = {Hengl, T. et al.},
  journal = {PLOS ONE},
  volume  = {10},
  number  = {6},
  pages   = {e0125814},
  year    = {2015},
  doi     = {10.1371/journal.pone.0125814}
}

@article{armand2018rmqs,
  title   = {The French Soil Quality Monitoring Network (RMQS): A tool for monitoring changes in soil properties at the national scale},
  author  = {Arrouays, Dominique and Saby, Nicolas P. A. and Walter, Christian and Lemercier, Beno{\^i}t and Schvartz, Christian},
  journal = {Soil Use and Management},
  volume  = {34},
  number  = {3},
  pages   = {303--313},
  year    = {2018},
  doi     = {10.1111/sum.12435}
}

@article{lark2012nsi,
  title   = {The National Soil Inventory of England and Wales: sampling design and monitoring of soil properties},
  author  = {Lark, R. M. and Bellamy, P. H. and Rawlins, B. G.},
  journal = {European Journal of Soil Science},
  volume  = {63},
  number  = {6},
  pages   = {887--898},
  year    = {2012},
  doi     = {10.1111/j.1365-2389.2012.01498.x}
}

@misc{ncss2023,
  author       = {{Soil Survey Staff}},
  title        = {National Cooperative Soil Survey Soil Characterization Database},
  year         = {2023},
  organization = {Natural Resources Conservation Service, United States Department of Agriculture},
  note         = {Available at https://ncsslabdatamart.sc.egov.usda.gov/}
}

@inproceedings{devlin2019bert,
  title={BERT: Pre-training of Deep Bidirectional Transformers for Language Understanding},
  author={Devlin, Jacob and Chang, Ming-Wei and Lee, Kenton and Toutanova, Kristina},
  booktitle={Proceedings of the 2019 Conference of the North American Chapter of the Association for Computational Linguistics (NAACL)},
  pages={4171--4186},
  year={2019}
}

@article{gorishniy2021revisiting,
  title={Revisiting Deep Learning Models for Tabular Data},
  author={Gorishniy, Yury and Rubachev, Ivan and Khrulkov, Valentin and Babenko, Artem},
  journal={Advances in Neural Information Processing Systems},
  volume={34},
  pages={18932--18943},
  year={2021}
}

@inproceedings{kendall2017uncertainties,
  title={What Uncertainties Do We Need in Bayesian Deep Learning for Computer Vision?},
  author={Kendall, Alex and Gal, Yarin},
  booktitle={Advances in Neural Information Processing Systems},
  volume={30},
  year={2017}
}

@article{jakubik2025terramind,
  title={TerraMind: Large-Scale Generative Multimodality for Earth Observation},
  author={Jakubik, Johannes and Yang, Felix and Blumenstiel, Benedikt and Scheurer, Erik and Sedona, Rocco and Maurogiovanni, Stefano and Bosmans, Jente and Dionelis, Nikolaos and Marsocci, Valerio and Kopp, Niklas and Ramachandran, Rahul and Fraccaro, Paolo and Brunschwiler, Thomas and Cavallaro, Gabriele and Bernabe-Moreno, Juan and Long{\'e}p{\'e}, Nicolas},
  journal={arXiv preprint arXiv:2504.11171},
  year={2025}
}

@article{pastorello2020fluxnet2015,
  title={The FLUXNET2015 dataset and the ONEFlux processing pipeline for eddy covariance data},
  author={Pastorello, Gilberto and Trotta, Carlo and Canfora, Eleonora and Chu, Housen and Christianson, Danielle and Cheah, You-Wei and Poindexter, Cristina and Chen, Jiquan and Elbashandy, Abdelrahman and Humphrey, Marty and others},
  journal={Scientific Data},
  volume={7},
  number={1},
  pages={225},
  year={2020},
  publisher={Nature Publishing Group},
  doi={10.1038/s41597-020-0534-3}
}

@article{orgiazzi2018lucas,
  title={LUCAS Soil, the largest expandable soil dataset for Europe: a review},
  author={Orgiazzi, Alberto and Ballabio, Cristiano and Panagos, Panos and Jones, Arwyn and Fern{\'a}ndez-Ugalde, Oihane},
  journal={European Journal of Soil Science},
  volume={69},
  number={1},
  pages={140--153},
  year={2018}
}

@article{johnson2023mimiciv,
  title={MIMIC-IV, a freely accessible electronic health record dataset},
  author={Johnson, Alistair E. W. and Bulgarelli, Lucas and Shen, Lu and Gayles, Alyssa and Shammout, Ahmad and Horng, Steven and Pollard, Tom J. and Hao, Shengpu and Moody, Benjamin and Gow, Brian and Lehman, Li-wei H. and Mark, Roger G. and Celi, Leo Anthony},
  journal={Scientific Data},
  volume={10},
  number={1},
  pages={1--16},
  year={2023},
  publisher={Nature Publishing Group}
}

@misc{nominatim,
  author = {{OpenStreetMap contributors}},
  title = {Nominatim},
  year = {2023},
  url = {https://nominatim.org/},
  note = {Accessed: 2026-04-13}
}

@misc{gadm,
  author = {{GADM contributors}},
  title = {GADM database of Global Administrative Areas},
  year = {2024},
  url = {https://gadm.org/},
  note = {Accessed: 2026-04-13}
}

@article{lee2025mirrams,
  title={MIRRAMS: Learning Robust Tabular Models under Unseen Missingness Shifts},
  author={Lee, Jihye and Kang, Minseo and Kim, Dongha},
  journal={arXiv preprint arXiv:2507.08280},
  year={2025}
}

@inproceedings{samad2024ifial,
  title={Imputation-free Learning of Tabular Data with Missing Values using Incremental Attention},
  author={Samad, Md and others},
  booktitle={ICLR Workshop / OpenReview preprint},
  year={2024}
}

@inproceedings{hegselmann2023tabllm,
  title={TabLLM: Few-shot Classification of Tabular Data with Large Language Models},
  author={Hegselmann, Stefan and Buendia, Alejandro and Lang, Hunter and Agrawal, Monica and Jiang, Xiaoyi and Sontag, David},
  booktitle={Proceedings of AISTATS},
  volume={206},
  pages={5549--5581},
  year={2023}
}

@article{jaitly2023serialization,
  title={Towards Better Serialization of Tabular Data for Few-shot Classification with Large Language Models},
  author={Jaitly, Sukriti and Shah, Tanay and Shugani, Ashish and Grewal, Razik Singh},
  journal={arXiv preprint arXiv:2312.12464},
  year={2023}
}

@article{valdenegro2025inputuncertainty,
  title={Can Bayesian Neural Networks Explicitly Model Input Uncertainty?},
  author={Valdenegro-Toro, Matias and others},
  journal={arXiv preprint arXiv:2501.08285},
  year={2025}
}

@article{buehler2024combining,
  title={Combining Input, Data and Model Uncertainty into a Single Neural Network},
  author={Buehler, Markus and others},
  journal={arXiv preprint arXiv:2406.18787},
  year={2024}
}

@inproceedings{huang2020tabtransformer,
  title={TabTransformer: Tabular Data Modeling Using Contextual Embeddings},
  author={Huang, Xin and others},
  booktitle={arXiv preprint arXiv:2012.06678},
  year={2020}
}

@inproceedings{arik2021tabnet,
  title={TabNet: Attentive Interpretable Tabular Learning},
  author={Arik, Sercan and Pfister, Tomas},
  booktitle={AAAI},
  year={2021}
}

@inproceedings{jaegle2021perceiver,
  title={Perceiver: General Perception with Iterative Attention},
  author={Jaegle, Andrew and others},
  booktitle={ICML},
  year={2021}
}

@misc{gemma2024,
  title={Gemma: Open Models Based on Gemini Research and Technology},
  author={Google DeepMind},
  year={2024}
}

@article{minasny2024soil,
  title={Soil science-informed machine learning},
  author={Minasny, Budiman and Bandai, Toshiyuki and Ghezzehei, Teamrat A and Huang, Yin-Chung and Ma, Yuxin and McBratney, Alex B and Ng, Wartini and Norouzi, Sarem and Padarian, Jose and Sharififar, Amin and others},
  journal={Geoderma},
  volume={452},
  pages={117094},
  year={2024},
  publisher={Elsevier}
}

@article{minasny2025machine,
  title={Machine learning and artificial intelligence applications in soil science},
  author={Minasny, Budiman and McBratney, Alex B},
  journal={European Journal of Soil Science},
  volume={76},
  number={2},
  pages={e70093},
  year={2025},
  publisher={Wiley Online Library}
}

@inproceedings{mandal2025irrisight,
  title = {{IRRISIGHT}: A Large-Scale Multimodal Dataset and Scalable Pipeline to Address Irrigation and Water Management in Agriculture},
  author = {Mandal, Nibir Chandra and Hoque, Oishee Bintey and Wilson, Mandy L. and Swarup, Samarth and Nouwakpo, Sayjro K. and Adiga, Abhijin and Marathe, Madhav},
  booktitle = {Advances in Neural Information Processing Systems},
  note = {Datasets and Benchmarks Track},
  year = {2025},
  url = {https://openreview.net/forum?id=SRP9tz3hYs}
}

@article{paudel2025cybench,
  title = {{CY-Bench}: A comprehensive benchmark dataset for sub-national crop yield forecasting},
  author = {Paudel, Dilli and Kallenberg, Michiel and Ofori-Ampofo, Stella and Baja, Hilmy and van Bree, Ron and Potze, Aike and Poudel, Pratishtha and Saleh, Abdelrahman and Anderson, Weston and von Bloh, Malte and Castellano, Andres and Ennaji, Oumnia and Hamed, Raed and Laudien, Rahel and Lee, Donghoon and Luna, Inti and Meroni, Michele and Mutuku, Janet Mumo and Mkuhlani, Siyabusa and Richetti, Jonathan and Ruane, Alex C. and Sahajpal, Ritvik and Shai, Guanyuan and Sitokonstantinou, Vasileios and de Souza N{\'o}ia J{\'u}nior, Rog{\'e}rio and Srivastava, Amit Kumar and Strong, Robert and Sweet, Lily-belle and Vojnovic, Petar and Athanasiadis, Ioannis N.},
  journal = {Earth System Science Data Discussions},
  year = {2025},
  doi = {10.5194/essd-2025-83},
  note = {Preprint, in review},
  url = {https://doi.org/10.5194/essd-2025-83}
}

@article{corcoran2026cycless,
  title = {A comprehensive {UK} crop yield dataset incorporating satellite, weather, and soil type information},
  author = {Corcoran, Evangeline and Bebber, Daniel P. and Curceac, Stelian and Efremova, Natalia and Lashkari, Azam and Mead, Andrew and Morris, Richard J. and Pywell, Richard F. and Redhead, John W. and Ahnert, Sebastian E.},
  journal = {Scientific Data},
  volume = {13},
  number = {1},
  pages = {491},
  year = {2026},
  doi = {10.1038/s41597-025-06528-x},
  url = {https://doi.org/10.1038/s41597-025-06528-x}
}

@article{su2021cropctnt,
  title = {A global dataset for crop production under conventional tillage and no tillage systems},
  author = {Su, Yang and Gabrielle, Benoit and Makowski, David},
  journal = {Scientific Data},
  volume = {8},
  number = {33},
  year = {2021},
  doi = {10.1038/s41597-021-00817-x},
  url = {https://doi.org/10.1038/s41597-021-00817-x}
}

@incollection{zhou2025agribench,
  title = {{AgriBench}: A Hierarchical Agriculture Benchmark for Multimodal Large Language Models},
  author = {Zhou, Yutong and Ryo, Masahiro},
  booktitle = {Computer Vision -- ECCV 2024 Workshops},
  editor = {Del Bue, Alessio and Canton, Cristian and Pont-Tuset, Jordi and Tommasi, Tatiana},
  series = {Lecture Notes in Computer Science},
  volume = {15625},
  pages = {207--223},
  publisher = {Springer},
  address = {Cham},
  year = {2025},
  doi = {10.1007/978-3-031-91835-3_14},
  url = {https://doi.org/10.1007/978-3-031-91835-3_14}
}

@misc{chiaburu2025soilnet,
  title = {{SoilNet}: A Multimodal Multitask Model for Hierarchical Classification of Soil Horizons},
  author = {Chiaburu, Teodor and Singh, Vipin and Hau{\ss}er, Frank and Bie{\ss}mann, Felix},
  year = {2025},
  eprint = {2508.03785},
  archivePrefix = {arXiv},
  primaryClass = {cs.LG},
  doi = {10.48550/arXiv.2508.03785},
  url = {https://arxiv.org/abs/2508.03785}
}

@inproceedings{gauba2025agmmu,
  title = {{AgMMU}: A Comprehensive Agricultural Multimodal Understanding Benchmark},
  author = {Gauba, Aruna and Pi, Irene and Man, Yunze and Pang, Ziqi and Adve, Vikram S. and Wang, Yu-Xiong},
  booktitle = {Advances in Neural Information Processing Systems},
  note = {Datasets and Benchmarks Track},
  year = {2025},
  url = {https://openreview.net/forum?id=MQPZPtv8GG}
}

@article{panagos2024euso,
  author  = {Panagos, Panos and Broothaerts, Nils and Ballabio, Cristiano and Orgiazzi, Alberto and De Rosa, Daniele and Borrelli, Pasquale and Liakos, Leonidas and Vieira, Diana and Van Eynde, Elise and Arias Navarro, Cristina and Breure, Timo and Fendrich, Arthur and K{\"o}ninger, Julia and Labouyrie, Maeva and Matthews, Francis and Muntwyler, Anna and Jimenez, Juan M. and Wojda, Piotr and Yunta, Felipe and Marechal, Anne and Sala, Serenella and Jones, Arwyn},
  title   = {How the {EU} Soil Observatory is providing solid science for healthy soils},
  journal = {European Journal of Soil Science},
  year    = {2024},
  volume  = {75},
  number  = {3},
  pages   = {e13507},
  doi     = {10.1111/ejss.13507}
}

@article{panagos2025soilmonitoring,
  author  = {Panagos, Panos and Jones, Arwyn and Lugato, Emanuele and Ballabio, Cristiano},
  title   = {A Soil Monitoring Law for Europe},
  journal = {Global Challenges},
  year    = {2025},
  volume  = {9},
  number  = {3},
  pages   = {2400336},
  doi     = {10.1002/gch2.202400336}
}

@article{panagos2022esdac,
  author  = {Panagos, Panos and Van Liedekerke, Marc and Borrelli, Pasquale and K{\"o}ninger, Julia and Ballabio, Cristiano and Orgiazzi, Alberto and Lugato, Emanuele and Liakos, Leonidas and Hervas, Javier and Jones, Arwyn and Montanarella, Luca},
  title   = {European Soil Data Centre 2.0: Soil data and knowledge in support of the {EU} policies},
  journal = {European Journal of Soil Science},
  year    = {2022},
  volume  = {73},
  number  = {6},
  pages   = {e13315},
  doi     = {10.1111/ejss.13315}
}

\end{document}